# Enhanced Urban Traffic Management Using CCTV Surveillance Videos and Multi-Source Data Current State Prediction and Frequent Episode Mining

Shaharyar Alam Ansari[1*], Mohammad Luqman[2], Prof. Aasim Zafar[3], Savir Ali[4]

[1*]School of Computer Science Engineering & Technology, Bennett University, Greater Noida, Uttar Pradesh, India, shaharyargd@gmail.com, 0000-0002-0469-253X

[2]Department of Computer Science, Aligarh Muslim University, Aligarh, Uttar Pradesh, India, luqman.geeky@gmail.com, 0000-0002-3564-8275

[3]Department of Computer Science, Aligarh Muslim University, Aligarh, Uttar Pradesh, India, prof.aasimzafar@gmail.com, 0000-0003-1331-014X

[4]Department of Computer Science, Aligarh Muslim University, Aligarh, Uttar Pradesh, India, savirali.cs@gmail.com, 0000-0001-6830-1027

**Abstract**

Rapid urbanization has intensified traffic congestion, environmental strain, and inefficiencies in transportation systems, creating an urgent need for intelligent and adaptive traffic management solutions. Conventional systems relying on static signals and manual monitoring are inadequate for the dynamic nature of modern traffic. This research aims to develop a unified framework that integrates CCTV surveillance videos with multi-source data descriptors to enhance real-time urban traffic prediction. The proposed methodology incorporates spatio-temporal feature fusion, Frequent Episode Mining for sequential traffic pattern discovery, and a hybrid LSTM–Transformer model for robust traffic state forecasting. The framework was evaluated on the CityFlowV2 dataset comprising 313,931 annotated bounding boxes across 46 cameras. It achieved a high prediction accuracy of 98.46%, with a macro precision of 0.9800, macro recall of 0.9839, and macro F1-score of 0.9819. FEM analysis revealed significant sequential patterns such as moderate–congested transitions with confidence levels exceeding 55%. The 46 sustained congestion alerts are system-generated, which shows practical value for proactive congestion management. This emphasizes the need for the incorporation of video stream analytics with data from multiple sources for the design of real-time, responsive, adaptable multi-level intelligent transportation systems, which makes urban mobility smarter and safer.

1. Introduction

Urban transport is facing a massive revolution as the world's cities are trying to find ways in which they can overcome challenges of congestion, environmental deterioration, and poor transport systems [1]. As the populations in cities are growing at a rapid pace, particularly in the developing world, traffic congestion is now a part and parcel of everyday activity with effects on economic productivity, commuter comfort, fuel usage, as well as living standards in general [2,3]. Conventional traffic control systems, typically supported by fixed settings and visual monitoring, cannot cope with the challenges and realities of modern-day traffic situations [4,5]. In search of a way to bridge gaps, the incorporation of information and communication systems, in general, and video surveillance systems, in particular, has been referred to as a potential answer [6,7].

CCTV cameras were once regarded as police equipment for monitoring and security purposes. At present, it is one of the most important components of the traffic management systems infrastructure [8]. Cameras provide a non-stop flow of video at intersections and along the major corridors, capturing spatial and temporal data related to traffic flows, incidents, and the density of traffic [9,10]. Image understanding with machine intelligence permits the extraction of structured data of some traffic parameters like counts, lane occupancy, and speed from the video recordings made by CCTV [11,12]. On the other hand, analysis dependent solely on CCTV constitutes one of the narrow perspectives and does not allow any reflection on what happens in the real world. Wet weather, the prevailing surface conditions, any event occurring, traffic lights being changed or not, are not the whole parent array of parameters intimately interacting with traffic situations [13,14]. The multi-source traffic condition dynamics can be better understood by incorporating GPS records with data from loop detectors, meteorological sensors, social media, and mobile apps [15-17].

Translating data from multidimensional and high-dimensional datasets into actionable insights has often proved difficult because of their various characteristics. In the case of the traffic flow, the datasets are temporal and sequential: what happens now comes to affect what can happen in the next moment [18,19]. Capturing such cause-and-effect sequences is what traditional time series analysis usually fails to do. For this purpose, FEM is utilized as an analysis tool [20,21]. Among others, FEM provides relevant insights into sequences of events in traffic flow, i.e., from medium flow to congestion to medium flow again [22-24].

This research provides solutions to the issues mentioned above through a robust framework in which video surveillance data is integrated and merged with other source data to derive current and future traffic. The framework minutes CCTV footage online while merging different camera streams in a time-ordered manner. It also implements spatio-temporal fusion with other traffic data to determine traffic dependency from one location to others. The FEM is employed to forecast sequences of events and traffic situations that are useful in the prediction of future traffic. Within this framework, context-aware hybrid models with deep learning architectures were built by merging LSTM networks with transformer-based attention structures so that the integrated systems capture all sequential trends and global city zone-level content. The purpose of this study is to develop an adaptive, scalable, and intelligent traffic prediction system integrating both CCTV surveillance and multi-source data analytics for real-time urban mobility management support. Key contributions of this work include:

- An integrated framework for predicting traffic that combines visual surveillance data with time and space characteristics.

- A novel Frequent Episode Mining method to discover critical traffic behaviour sequences.

- A hybrid LSTM–Transformer architecture tailored for traffic state forecasting.

- Preprocessing modules for time-stamp alignment, vehicle count smoothing, and anomaly normalization.

- An open-source vehicle counting tool for structured data generation from video streams.

- Evaluation of the model on a real-world benchmark dataset shows high accuracy in classifying traffic states and detecting sustained congestion.

The research not only addresses the technical design but also evaluates the effectiveness of the proposed approach using real-world datasets and empirical case studies. In doing so, it contributes to the ongoing discourse on intelligent transportation systems and their role in shaping smarter, safer, and more adaptive cities.

The structure of the paper is as follows: in Section 1, the introduction of the research topic is provided in detail. In section 2, the previous studies that are related to the urban traffic management using CCTV surveillance videos and multi-source data, current state prediction, and frequent episode mining are discussed. Further, the research methodology of this research is provided in detail, along with the dataset and utilized techniques. Section 4 provides the results and discussion of the study in detail. Conclusion and future directions of the research are provided in the final section 5.

## 2. Literature Review

This literature review summarizes the existing literature on the problem of multi-source data mining and prediction with frequent episode mining.

Ashkanani et al. (2025) [25] proposed Adaptive traffic control as an optimization model using machine learning methods on video surveillance footage to modify signal timings according to the real-time traffic volume and conditions. Experimental results indicated that You Only Look Once (YOLOv11) attained a higher level of efficiency and an overall accuracy rating of about 95.1% compared to its predecessors. In the same year, Chaudhary et al. (2025) [26] introduced a new system for managing traffic congestion in smart cities with Dynamic Tabu Search-based Gated Recurrent Units (DTS-GRU). Based on the trial results, the proposed solution was beneficial for F1-score (92.90%), recall (94.60%), accuracy (93.10%), and precision (94.05%) benchmarks. Similarly, Charef et al. (2025) [27] showcased a mobile application designed specifically to revolutionize the methods employed in the collection of urban traffic data. Smartphones were used as cameras in capturing traffic scenes. At the same time, the program was built on the reliable mobile version of YOLOv8, which achieved 93% accuracy in real-time traffic monitoring and vehicle counting.

Previously, Zahra et al. (2024) [28] designed an architecture that incorporated a video surveillance system based on deep learning, which encoded a video frame by compressing it after important areas had been cropped without any form of information loss. For two benchmark datasets, the technique achieved a PSNR of 5.35 bd and segmentation accuracy of 92% and 96% SR-based secondary-augmented segmentation precision. Meanwhile, Li et al. (2024) [29] developed the first multi-level Tilt Camera-Controlled (TTC-X) surveillance system, which introduced dynamic and autonomous traffic monitoring and control in urban networks. As per the experimental results, TTC-X successfully predicted link-level traffic conditions with the best Mean Absolute Error (MAE) of below 1.25 vehicles per hour, routed under an unanticipated complete lane-closure strategy, and over 60% capture rate for all cars at the network level. Moreover, Devadhas Sujakumari and Dassan (2023) [30] developed a generic ontology for any IoT Security scenario known as the IoT Security Threat Ontology (IoTSTO), which was used to specify the many facets of IoT security threats and to generate inference rules for analyzing the threats. The results of this research addressed the semantic heterogeneity of knowledge gleaned from several sources. They completed the network of publicly available knowledge repositories dedicated to the Internet of Things security.

The usefulness of frequent episode mining to uncover temporal relationships within collections of event types, even when they are disordered, was discussed by Qin et al. (2022) in [31]. To enhance episode frequency detection while ensuring privacy, a sample-based perturbation technique was proposed. The effectiveness and efficiency of this method were evaluated through real-life datasets. However, Jiang et al. (2022) [32] presented a holographic traffic signal control system incorporating multi-source data fusion to the traffic data collected from different sensors. The results showed that the average time spent waiting at a single intersection scaled down by 25.03%, while the parking delay average decreased by 61.51%. In the past, Li et al. (2022) [33] used a combination of a Gated Recurrent Unit (GRU) and an LSTM to aggregate nodes in a subgraph, as well as the edge weight and information transmission method that was optimal for the data in the subgraph. This study's proposed strategy for predicting stock market volatility outperformed the dimensional reduction index by 16.64%. It outperformed previous methods for fusing and predicting heterogeneous data by 14.48%, all while utilizing the same model. On the other hand, Zeng et al. (2021) [34] suggested the new DQ-DTA (Data-Driven Quasi-Dynamic Traffic Assignment) model. Time-varying travel demand was estimated in this model using toll station records. Test findings show that the suggested DQ-DTA model's accuracy is roughly 6% more than the selected STA models.

Despite notable advancements in multi-source traffic data mining and real-time congestion prediction, three key research gaps persist in existing literature. Unlike YOLOv11, which uses camera feeds in spatio-temporal isolation and TTC-X, which lacks robust temporal fusion, much work remains to be done in spatio-temporal fusion [25,29]. Second, existing solutions often prioritize detection accuracy but fall short in capturing long-term temporal dependencies and sequential evolution of traffic states, as seen in [27,28,31]. Unlike [32], which focuses on holographic systems, much of the work done in the field focuses on the lack of real-time congestion alerts, actionable congestion alerts, and systems devoid of interpretability. The next section covers the dataset, techniques, and the development of advanced urban traffic management systems to close these gaps using multi-source data and CCTV surveillance footage.

3. **Research Methodology**

This section provides the dataset and proposed methodology for the advanced urban traffic management systems that fuse CCTV surveillance videos and multi-source data. The very first step of the analysis is data initialization and involves the extraction, alignment, and synchronization of vehicle counts (NoVD), anomaly tags (AoVD), and time-stamps (TiS) to a reference time. These inputs are aligned across multiple video sources to form a unified event stream. Spatio-temporal feature fusion captures inter-location dependencies, while FEM identifies recurring patterns such as high traffic followed by anomalies. These insights support a traffic state prediction module, which forecasts congestion and risks. Real-time performance is ensured through data flow maintenance, and the system is evaluated using metrics like RMSE, MAE, precision, recall, and episode confidence.

### 3.1 Dataset

The Track 1 dataset [35], known as CityFlowV2, comprises approximately 3.58 hours (215 minutes) of video captured by 46 high-resolution (≥960p), predominantly 10 FPS cameras, covering 16 intersections across a mid-sized U.S. city, with the furthest simultaneous cameras spaced about 4 km apart. The dataset is organized into six scenarios, three for training, two for validation, and one for testing, and includes 313,931 annotated bounding boxes corresponding to 880 unique vehicle identities, each annotated only if the vehicle appears in at least two different camera views. Each video provides time-offset metadata for synchronization, and frames are indexed using a 0-based coordinate system, with alphanumeric video IDs assigned in ascending order. Samples of the dataset are shown in Figure 1.

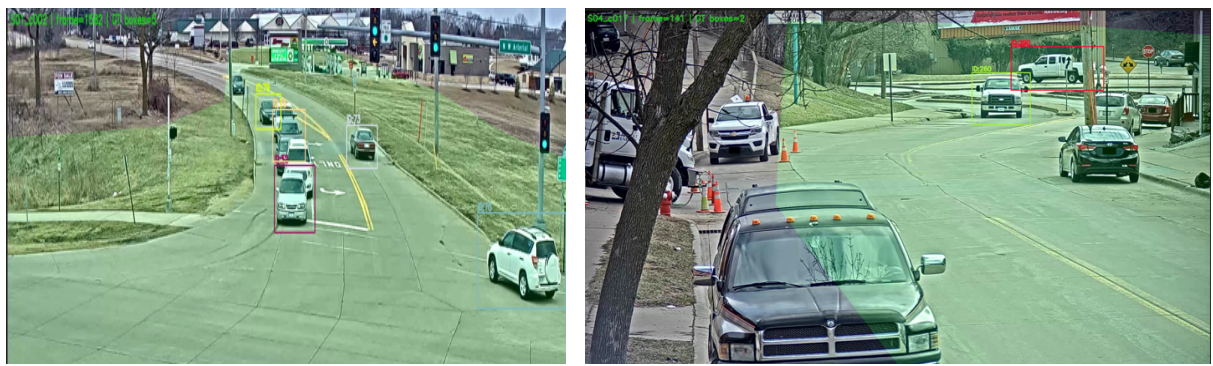

**Figure 1: Dataset Samples**

### 3.2 Data Preprocessing

The data preprocessing step plays a crucial role in preparing raw video descriptors for structured analysis and modeling [36]. The methodology it involves synchronizing time-stamps (TiS) across multiple CCTV video sources to ensure temporal consistency, smoothing vehicle count sequences (NoVD) to reduce noise and fluctuations caused by detection errors, and normalizing anomaly tags (ANoV) into a unified format for consistent interpretation across different traffic scenarios. This step ensures that all input data, regardless of source location or recording conditions, is standardized and aligned for accurate feature extraction, spatio-temporal fusion, and frequent episode mining.

### 3.3 Feature Extraction

In the proposed methodology, feature extraction [37] refers to getting meaningful frame-level descriptors from preprocessed video data that give away dynamic information about the state of urban traffic. For each frame, key features such as the number of vehicles (NoVD), anomaly tags (ANoV), and time information (TiS), and a derived label of traffic class (with predefined thresholds), such as free-flow, moderate, or congested, are extracted. These features are then structured into a common format that allows consistent analysis across several locations and time windows. This step permits converting raw observations into analyzable traffic event sequences, which become the basis for downstream processes, from spatio-temporal fusion to frequent episode mining and, finally, to traffic state prediction.

### 3.4 Time-Space Sequence Alignment

Time-Space Sequence Alignment means the superimposition and arrangement of time-stamped data from multiple sources located at different spatial locations (a few CCTV cameras in various locations, for example) on a single coherent chronological axis [38]. It emphasizes those events that were recorded in different places at different times--to ensure comparability, thereby allowing for joint consideration of the temporal pattern and spatial dependency in the traffic behaviour being studied [39]. This step, in the methodology proposed herein, involves aligning the event streams output by multiple CCTV surveillance videos, with respect to their temporal identifiers (TiS) and spatial identifiers (Video_ID). This step aligns the extracted features, such as NoVD, ANoV, and derived traffic classes, into a unified, temporally ordered sequence across all monitored locations. This alignment is critical for subsequent stages such as feature fusion, episode mining, and predictive modeling. Time-Space Alignment aims to construct a unified sequence:

$$S = \bigcup_{i=1}^{N} E_i \ sorted \ by \ t_{ij}$$

Such that:

- All events are merged into a global timeline S

- Events retain spatial identifiers $l_i$

- Temporal order across all locations is preserved

### 3.5 Spatio-Temporal Feature Fusion

Spatio-Temporal Feature Fusion [40] is the fusion of learned features from different spatial points and different consecutive time intervals in order to build a joint representation capturing both spatial and temporal dynamics. With spatio-temporal feature fusion, it is capable of capturing interdependencies in different regions and time intervals and facilitating accuracy in traffic state estimation and behavioral patterns extraction [41,42]. The joint representation in all N regions and the time interval T period is:

$$F_{fused} = F\left(\bigcup_{i=1}^{N}\bigcup_{t=1}^{T}F_i(t)\right)$$

Here, F(·) is the spatiotemporal fusion function, also elucidated as temporal stacking combined with spatial relation modeling. As per the proposed method, the second fusion step, i.e., spatio-temporal feature fusion, acts upon the extracted traffic features, vehicle count (NoVD), anomaly tags (ANoV), and traffic class labels from different locations of CCTVs between given time frames, to form the generalized representation of urban traffic dynamics. This helps in capturing spatial dependency, such as congestion propagation between intersections, and the temporal patterns, e.g., periodic buildup of congestion or sudden anomalies. In fact, this fusion facilitates the system in comprehending contextual traffic behavior occurring within different city zones and at other time stages, thereby improving further traffic management decisions.

### 3.6 Frequent Episode Mining

FEM [43] is a temporal pattern mining technique used to discover sequences of events that frequently occur in a specific time order within a defined time window [44]. In the proposed methodology, the Frequent Episode Mining (FEM) step identifies commonly occurring sequences of traffic events, such as high vehicle density followed by anomalies like wrong-side driving across synchronized video streams. After aligning and fusing features like vehicle count (NoVD), anomaly tags (ANoV), and time-stamps (TiS) into structured event sequences, FEM is applied using a sliding time window and support threshold to detect significant temporal patterns. These episodes help reveal critical cause-and-effect chains and behavioral trends in urban traffic, which are essential for understanding congestion dynamics and supporting proactive traffic state prediction.

**Pseudocode: Frequent Episode Mining (FEM) Using TCS-Tree**

Input:

    EventStream S = [e$_1$, e$_2$, ..., e$_n$] where each e$_i$ = (Time, Location, NoVD, ANoV, Class)

    WindowSize W

    MinSupport θ

Output:

    Set of Frequent Episodes E

Procedure FEM_TCS_Tree(S, W, θ):

-Initialize TCS_Tree T ← Empty

-For i = 1 to n:

  CurrentTime ← S[i].Time

  EpisodeWindow ← Extract all events e_j where (CurrentTime - S[j].Time) ≤ W

-For each subsequence seq in EpisodeWindow:

  If seq not in T:

   Insert seq into T with count = 1

  Else:

   T[seq].count += 1

-FrequentEpisodes ← {}

-For each seq in T:

  If T[seq].count ≥ θ:

   Add seq to FrequentEpisodes

-Return FrequentEpisodes

### 3.7 LSTM-Transfer Hybrid Model

The **LSTM-Transformer hybrid model** combines the sequential learning ability of Long Short-Term Memory (LSTM) networks with the parallel attention mechanism of Transformers to capture both long-range dependencies and context relevance in time-series data [45]. It leverages LSTM for temporal encoding and Transformer layers for contextual refinement across sequences [46]. In the proposed methodology, the LSTM-Transformer hybrid model is employed in the traffic state prediction module to learn spatio-temporal dependencies from fused traffic features effectively. First, LSTM layers process the time-aligned input sequence $X = [x_1, x_2, \ldots, x_T]$, where each xt represents a frame-level fused vector of vehicle count, anomaly tag, and location identifier. The LSTM computes hidden states such as:

$$h_t = LSTM(x_t, h_{t-1})$$

These hidden states $H = [h_1, h_2, \ldots, h_T]$ are then passed to a Transformer encoder, where the attention mechanism enhances contextual understanding:

$$AAttention(Q, K, V) = softmax\left(\frac{QK^\top}{\sqrt{d_k}}\right)V$$

Here, Q, K, and V are the query, key, and value matrices derived from H, and dk is the dimension of keys. The model outputs a context-aware representation that supports accurate prediction of traffic states, such as congestion level or anomaly risk, by considering both temporal sequence patterns and inter-frame contextual relationships.

After traffic prediction using a hybrid model, the final stage of the proposed methodology is Data Flow Maintenance, which ensures the continuous and synchronized processing of traffic data streams by simulating real-time conditions using rolling buffers, timestamp-based sequencing, and dynamic memory handling to support timely updates, anomaly alerts, and model input consistency. Following this, Performance Evaluation Metrics are applied to assess the system's effectiveness, using statistical measures such as RMSE and MAE for vehicle count prediction, precision, recall, and F1-score for anomaly detection accuracy, and support and confidence values for the quality of mined episodes by ensuring the model's reliability, responsiveness, and practical utility in urban traffic management.

**3.8  Proposed Methodology:**

In this section, the architecture of the proposed methodology is provided in Figure 2.

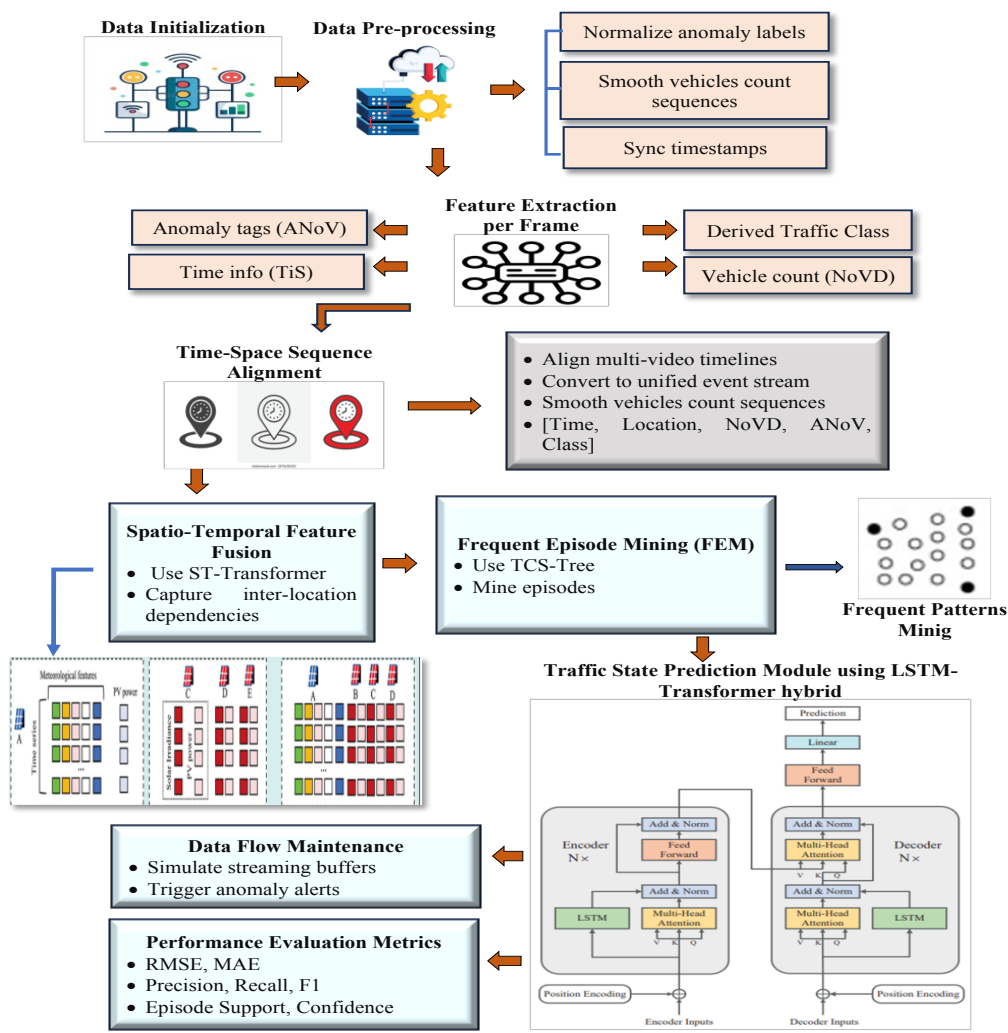

**Figure 2. Flow chart of the proposed methodology**

### 3.9 Proposed Algorithm:

**Start**

// **Phase 1: Data Initialization**

Let D = {$D_1, D_2, ..., D_n$}  // Input data sources (e.g., video feeds, traffic sensors)

**For each Di in D:**

   If Di is valid:

      Load Di

   Else:

      Log "Data source Di missing or corrupted."

// **Phase 2: Data Preprocessing**

**For each Di in D:**

   Normalize anomaly labels in Di

   Smooth vehicle count sequences in Di

   Sync time-stamps across all Di

// **Phase 3: Feature Extraction per Frame**

Let F = ∅  // Feature list

**For each frame f in Di:**

   Extract Anomaly Tags (ANoV)

   Extract Time Information (TiS)

   Extract Vehicle Count (NoVD)

   Derive Traffic Class (Class)

   F ← F ∪ {Time, Location, NoVD, ANoV, Class}

// **Phase 4: Time-Space Sequence Alignment**

Let A = ∅

Align multi-video timelines

Convert to unified event stream

Smooth vehicle count sequences

A ← A ∪ {Aligned Time-Space Sequences}

**// Phase 5: Spatio-Temporal Feature Fusion**

**Let ST = ∅**

Use Spatio-Temporal Transformer (ST-Transformer)

**For each time-stamp t:**

   Capture inter-location dependencies

   ST ← ST ∪ Fusion(A[t])

**// Phase 6: Frequent Episode Mining (FEM)**

Let E = ∅

Use TCS-Tree structure

**For each window W in ST:**

   Mine frequent episodes:

     [High NoVD → Anomaly → Wrong Side]

   E ← E ∪ Mined_Episodes(W)

**// Phase 7: Traffic State Prediction using LSTM-Transformer Hybrid**

Let C = ∅   // Predicted states

For each episode e in E:

   **Predict state using a hybrid model:**

     Encoder (LSTM + Transformer)

     Decoder (Dense Layer)

   C ← C ∪ Predicted_State(e)

**// Phase 7 (a): Data Flow Maintenance**

Let T = ∅

Simulate streaming buffers

**For each predicted state c in C:**

   Trigger anomaly alerts if necessary

   T ← T ∪ Updated_Flow(c)

**// Phase 7 (b): Performance Evaluation**

Let TP = 0, TN = 0, FP = 0, FN = 0

**For each prediction p in C:**

  Compare p with ground truth

  Update TP, TN, FP, FN accordingly

**Compute:**

  - **Precision** = TP / (TP + FP)

  - **Recall** = TP / (TP + FN)

  - **F1 Score** = 2 * (Precision * Recall) / (Precision + Recall)

  - Episode Support and Confidence

**End**

## 4. Result Analysis

The implementation and results obtained from the proposed methodology are presented in this section. Several key aspects were analyzed and evaluated to assess the performance of the model.

The Track 1 dataset, which captures synchronized video streams from multiple cameras across subsets S01, S02, and S03, is shown in Table 1. Each entry logs key metadata such as camera ID, frame rate, resolution, total frames, and duration in seconds, which ensures consistency and temporal alignment. Most cameras recorded at 10 fps with 1920×1080 resolution, supporting reliable downstream processing. For example, S01 captured 1,955 frames over 195.5 seconds, while S03, under subset S02, recorded at a slightly lower 9.67 fps. This structured and uniform dataset enables seamless integration into later phases like feature extraction, anomaly detection, and frequent episode mining, laying a solid foundation for spatio-temporal traffic analysis.

Table 1: Synchronized video streams from multiple cameras across subsets S01, S02, and S03

| split | subset | cam | fps | frames | width | height | duration_sec |
|---|---|---|---|---|---|---|---|
| train | S01 | S01 | 10.0 | 1955 | 1920 | 1080 | 195.5 |
| train | S01 | c004 | 10.0 | 2110 | 1920 | 1080 | 199.6 |
| train | S01 | S03 | 10.0 | 2110 | 1920 | 1080 | 211.0 |
| train | S01 | S04 | 10.0 | 2140 | 1920 | 1080 | 214.1 |
| train | S01 | S05 | 10.0 | 2420 | 1920 | 1080 | 242.2 |
| train | S01 | S01 | 10.0 | 2340 | 1920 | 1080 | 227.9 |

| train | S02 | S03 | 9.67 | 2110 | 1920 | 1080 | 211.0 |
| train | S02 | S04 | 10.0 | 2115 | 1920 | 1080 | 214.5 |
| train | S02 | S05 | 10.0 | 2200 | 1920 | 1080 | 242.5 |
| train | S03 | S03 | 10.0 | 2332 | 1920 | 1080 | 233.2 |

The dataset is summarized in Table 2, which captures three key training subsets, i.e., S01, S02, and S03, by highlighting their scale and data readiness. Subset S01 recorded 10,281 frames at an average of 10 fps, spanning 17.14 minutes, with annotations available for 5 cameras across ground truth, detections, ROIs, and calibrations. S02 covered 13,517 frames at 9.67 fps, lasting 23.33 minutes, with complete metadata for 6 cameras. The largest in annotation density, S03 logged 10,780 frames over 17.97 minutes, with 25 fully annotated camera feeds. This consolidated view reflects the dataset's balance in temporal length, frame consistency, and comprehensive annotation coverage, making it well-suited for robust spatio-temporal traffic learning and analysis.

Table 2: Summary of Dataset

| Split | Subset | Total frames | Avg fps | Total duration min | With gt | With det | With roi | With calib |
|---|---|---|---|---|---|---|---|---|
| train | S01 | 10281 | 10.0 | 17.14 | 5 | 5 | 5 | 5 |
| train | S02 | 13517 | 9.67 | 23.33 | 6 | 6 | 6 | 6 |
| train | S03 | 10780 | 10.0 | 17.97 | 25 | 25 | 25 | 25 |

Figure 3 illustrates the vehicle count trend per frame for a specific camera (labeled as $D_{10}$) as part of the data preprocessing phase in the proposed methodology. The dashed blue line represents the raw vehicle count, which fluctuates sharply due to frame-wise detection noise and natural traffic variability. To improve the reliability of downstream traffic state analysis, a smoothing technique is applied, shown by the solid orange line, which effectively filters out sporadic spikes and captures the underlying traffic flow pattern more consistently. This smoothed signal provides a clearer temporal trajectory of vehicle movement, enabling more stable feature extraction and episode detection in subsequent stages. The visualization confirms the importance of signal smoothing during preprocessing to reduce noise without losing important traffic dynamics.

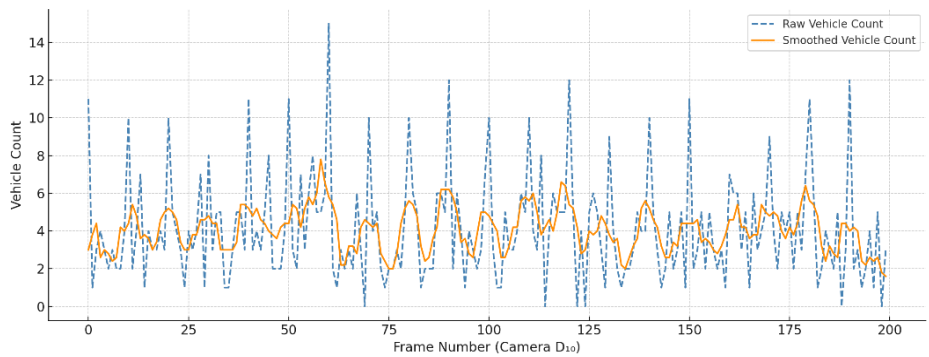

**Figure 3: Sample vehicle count before vs after smoothing**

Figure 4 shows the per-camera distribution of traffic conditions, i.e., congested, free-flow, and moderate, for the first 10 cameras used in the dataset. This classification outcome is obtained after applying feature extraction, where frame-wise vehicle counts, time-stamp alignment, and anomaly tags were used to assign traffic classes based on percentile thresholds. Each color-coded bar represents the number of frames labeled under a specific traffic class per camera. A clear trend emerges where the moderate condition (orange bars) dominates across nearly all cameras, especially S03/c011, S03/c013, and S03/c014, which exhibit the highest counts of moderate traffic frames. In contrast, free-flow (green) and congested (red) states show relatively balanced but lower frequencies. The visualization highlights spatial variation in traffic patterns across locations, and the dominance of moderate traffic frames supports the stability of the dataset for training predictive models and mining frequent traffic episodes in subsequent steps.

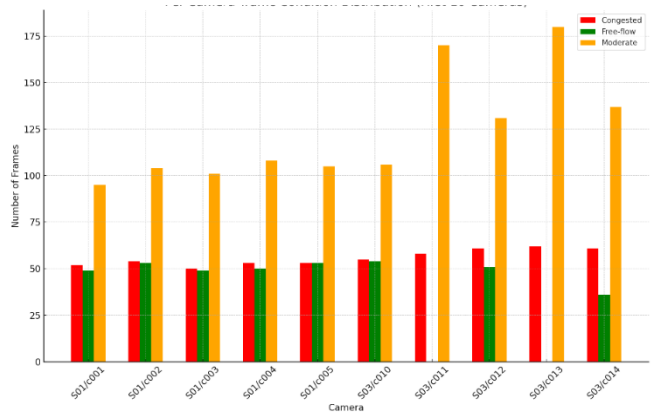

**Figure 4: Per-camera distribution of traffic conditions**

Figure 5 illustrates how traffic conditions that are congested, free-flow, and moderate change over time in 5-second intervals. Moderate traffic (blue) consistently occupies a large portion of the timeline, reflecting its dominance across most time bins. Free-flow (green) shows notable peaks, especially between 100 and 180 seconds, while congested conditions (orange) appear more frequently at the start and end. This temporal composition highlights how traffic conditions dynamically shift over time and reinforces the necessity of frequent episode mining and time-aware modeling in the proposed methodology. By aggregating class labels into time bins, the visualization effectively captures temporal trends in traffic states and offers a clearer picture of congestion cycles and transitional flow behaviors across the monitored segment.

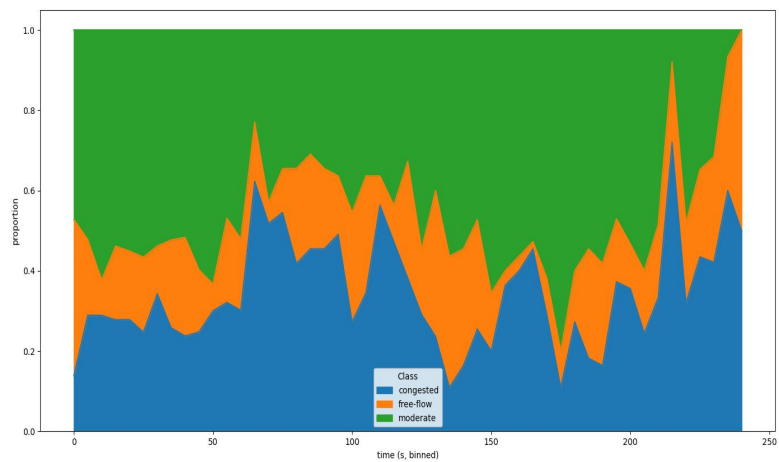

**Figure 5: Class Composition over Time**

Figure 6 illustrates the frame-wise change in Number of Vehicles Detected (Δ NoVD), a key component in Spatio-Temporal Feature Fusion. The x-axis represents the change in vehicle count between consecutive frames, while the y-axis indicates the number of frames that exhibit each level of change. The sharp central spike at Δ NoVD ≈ 0 shows that in most frames, the number of vehicles remains stable over time, suggesting gradual movement patterns and temporal continuity in the traffic stream. Smaller bars on either side represent minor increases or decreases in vehicle count, while extreme variations are rare. This stable delta pattern supports the use of temporal smoothing and fusion techniques, that is, Spatio-Temporal Transformers, as it confirms that traffic behavior evolves progressively rather than abruptly, making it suitable for learning temporal dependencies and enhancing prediction accuracy.

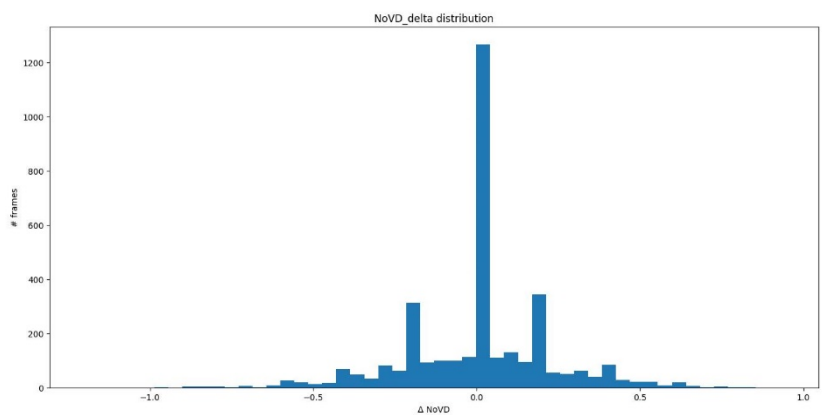

**Figure 6: NoVD delta distribution**

The application of the FEM algorithm on the spatio-temporal traffic dataset revealed ten significant k=5 event sequences, extracted using a TCS-Tree-based structure with sliding windows and a set support threshold. The most frequent pattern, "moderate → A0 → moderate → A0 → moderate → A0 → moderate", achieved a support of 0.03787 and a confidence of 46.34%, indicating recurring moderate traffic oscillation at intersection A0 as shown in Table 3. Another pattern, "free-flow → A0 → moderate → A0 → moderate", showed a high confidence of 78.35%, typically aligning with off-peak flow transitions. Episodes involving moderate–congested transitions also ranked high in frequency, emphasizing dynamic congestion onset. These temporal patterns serve as causal cues, enhancing the learning of sequential dependencies in the subsequent LSTM-Transformer-based traffic prediction module.

**Table 3: Temporal Patterns of the dataset using FEM**

| Rank | Episode (k=5) Preview | Length | Count | Support | Confidence |
|---|---|---|---|---|---|
| 0 | moderate → A0 → moderate → A0 → moderate → A0 → mode... | 5 | 133 | 0.037870 | 0.463415 |
| 1 | moderate → A0 → moderate → A0 → moderate → A0 → mode... | 5 | 121 | 0.034453 | 0.421063 |
| 2 | moderate → A0 → moderate → A0 → moderate → A0 → cong... | 5 | 93 | 0.026481 | 0.505435 |
| 3 | congested → A0 → moderate → A0 → moderate → A0 → mod... | 5 | 78 | 0.022210 | 0.503226 |
| 4 | free-flow → A0 → moderate → A0 → moderate → A0 → mod... | 5 | 76 | 0.021640 | 0.783505 |
| 5 | moderate → A0 → moderate → A0 → congested → A0 → mod... | 5 | 71 | 0.020216 | 0.449367 |
| 6 | moderate → A0 → congested → A0 → moderate → A0 → mod... | 5 | 66 | 0.018793 | 0.554622 |
| 7 | congested → A0 → moderate → A0 → moderate → A0 → mod... | 5 | 56 | 0.015945 | 0.368456 |
| 8 | moderate → A0 → congested → A0 → moderate → A0 → mod... | 5 | 53 | 0.015091 | 0.595506 |
| 9 | moderate → A0 → free-flow → A0 → moderate → A0 → mod... | 5 | 50 | 0.014237 | 0.649351 |

Following the frequent episode mining step, the LSTM-Transformer hybrid model is trained on the mined spatio-temporal patterns to learn temporal traffic dependencies. This model serves as the core of the Traffic State Prediction module, enabling accurate classification of evolving traffic conditions across time.

Following the LSTM-Transformer-based prediction, the confusion matrix (see Figure 7) validates the model's effectiveness across traffic classes. It shows strong accuracy with 114 free-flow, 432 moderate, and 475 congested frames correctly classified. Minimal confusion is observed, mainly between adjacent states, confirming the model's ability to distinguish traffic conditions with high reliability.

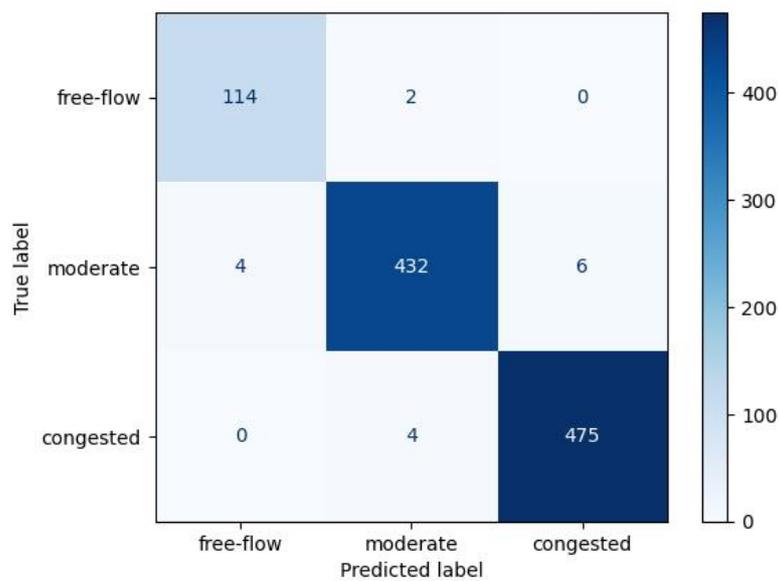

**Figure 7: Confusion Matrix**

The training progression of the hybrid LSTM-Transformer model is illustrated through accuracy and loss curves across 20 epochs, as shown in Figure 8 (a) and 8 (b), respectively. As shown in the first plot, the model's training accuracy sharply rises from approximately 58% in the first epoch to above 90% by the third epoch, ultimately reaching around 99% by the 20th epoch. The validation accuracy also follows a similar trajectory, starting near 86% and maintaining a stable performance close to 97% in the later epochs, indicating minimal overfitting. In parallel, the second plot reveals a steep decline in training loss from 0.85 to nearly 0.05 within the first five epochs, followed by a consistent downward trend that stabilizes below 0.05. Validation loss mirrors this behavior, decreasing rapidly from 0.45 to 0.07 and remaining stable thereafter. These patterns confirm the model's strong learning capability and generalization to unseen data, making it well-suited for reliable traffic state prediction based on the mined frequent event sequences.

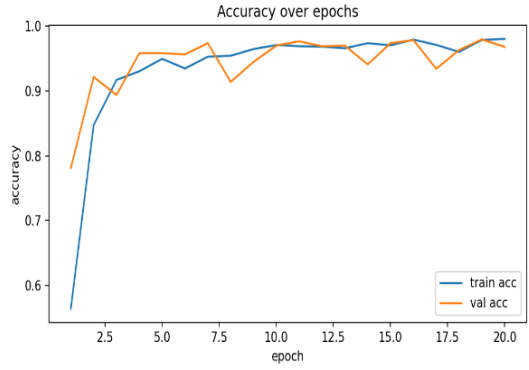 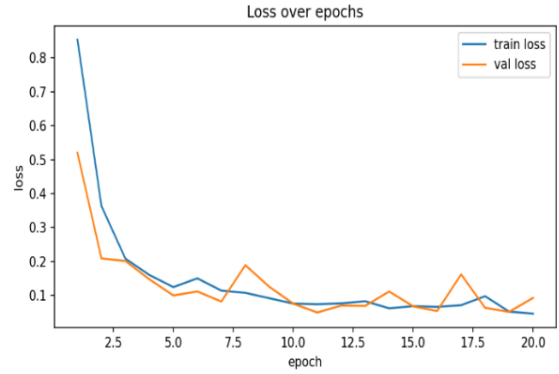

(a) (b)

**Figure 8: (a) Accuracy and (b) Loss of the hybrid model over Epochs**

Following the training and evaluation of the LSTM-Transformer model, the traffic state prediction framework enters its final stage of data flow maintenance and alert generation based on predicted congestion levels. The provided table captures instances where sustained congestion alerts were triggered across different cameras. These alerts are raised when the current vehicle count (NoVD) remains consistently above the defined threshold (Thr high) for a specified buffer length of 50 frames and 12 consecutive congested states as shown in Table 4. For example, at time 93.7 seconds, camera S01/c003 registered a current count of 6 with a buffer mean of 6.58, surpassing the threshold of 6.0, thus prompting an alert. This real-time decision-making logic enables timely congestion detection, ensuring that the system can proactively assist in traffic management by flagging hotspots with ongoing vehicular buildup. Overall, 46 alerts have been debounced, emphasizing prolonged congestion across various camera feeds.

**Table 4: Data flow maintenance and alert generation**

| Time (sec) | camera | NoVD current | Novd buffer mean | Thr high | Buffer len | Consec cong | Last class | status |
|---|---|---|---|---|---|---|---|---|
| 93.7 | S01/c003 | 6 | 6.58 | 6.0 | 50 | 12 | congested | ALERT: Sustained Congestion |
| 96.0 | S01/c001 | 4 | 3.36 | 6.0 | 50 | 12 | congested | ALERT: Sustained Congestion |
| 106.9 | S01/c004 | 5 | 4.36 | 3.0 | 50 | 12 | congested | ALERT: Sustained Congestion |
| 110.4 | S01/c004 | 5 | 4.33 | 4.0 | 50 | 12 | congested | ALERT: Sustained Congestion |
| 114.3 | S03/c014 | 3 | 2.62 | 6.0 | 50 | 12 | congested | ALERT: Sustained Congestion |
| 158.3 | S01/c002 | 4 | 4.82 | 4.4 | 50 | 12 | congested | ALERT: Sustained Congestion |
| 161.3 | S01/c002 | 4 | 3.54 | 4.0 | 50 | 12 | congested | ALERT: Sustained Congestion |
| 191.4 | S01/c010 | 2 | 3.50 | 2.0 | 50 | 12 | congested | ALERT: Sustained Congestion |
| 197.1 | S01/c002 | 5 | 4.55 | 5.0 | 50 | 12 | congested | ALERT: Sustained Congestion |
| 201.5 | S03/c014 | 3 | 2.82 | 6.0 | 50 | 12 | congested | ALERT: Sustained Congestion |

The final performance evaluation of the proposed traffic prediction framework demonstrates its strong generalization and balanced classification ability. The model achieved an impressive test accuracy of 98.46%, along with a macro precision of 0.9800, macro recall of 0.9839, and macro F1-score of 0.9819, indicating that it performed consistently well across all three traffic classes: free-flow, moderate, and congested, as shown in Figure 9.

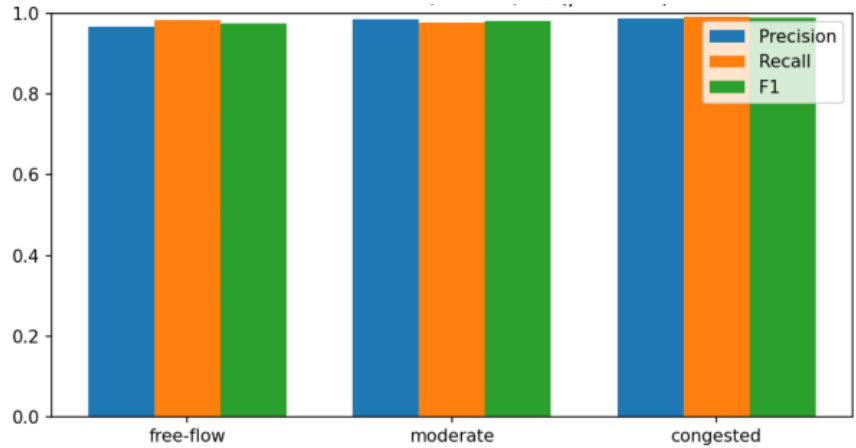

**Figure 9: Performance Analysis**

These high and closely aligned metrics validate the effectiveness of the Spatio-Temporal Feature Fusion combined with the LSTM-Transformer model, ensuring robust detection and classification of varying traffic states in real-world urban scenarios.

## 4.1 Comparative Analysis

Based on the comparative performance analysis of various methods used in multi-camera vehicle detection and tracking, the proposed LSTM Transfer Learning-based hybrid model demonstrates a significantly superior outcome in terms of F1-Score. The study includes comparisons with five other state-of-the-art techniques from previous studies, as shown in Table 5. Yao et al. (2022) [47] used R-CNN and YOLO to achieve an F1-score of 83.71 percent, focusing on city-scale multi-camera tracking with space-time-appearance features. Li et al. (2022) [48] implemented YOLOv5 for vehicle detection and achieved an improved score of 84.37 percent. Tang et al. (2019) [49] employed a technique combining temporal context with SSD, reaching an F1 score of 79.7 percent. Hsu et al. (2022) [50] proposed the SCLM approach, which yielded a significantly lower F1-score of 55.56 percent, indicating limited detection accuracy under their experimental conditions. Lu et al. (2023) [51] introduced a Box-Grained Matching method that performed comparatively better with an F1-score of 84.91 percent. In contrast, the proposed model combining LSTM with a transfer learning strategy delivered a remarkably high F1-Score of 98.19 percent, highlighting its robustness and efficiency in capturing temporal dependencies and domain adaptation for multi-camera environments.

**Table 5: Comparative Analysis**

| Authors | Method | F1-Score |
|---|---|---|
| **Yao et al. [2022] (47)** | R-CNN, YOLO | 83.71% |
| **Li et al. [2022] (48)** | YOLOv5 | 84.37% |

| Tang et al. [2019] (49) | TC+SSD | 79.7% |
| Hsu et al. [2022] (50) | SCLM | 55.56% |
| Lu et al. [2023] (51) | Box-Grained Matching method | 84.91% |
| **Proposed** | **LSTM-Transfer Hybrid Model** | **98.19%** |

This considerable improvement reflects the practical significance of the proposed approach in enhancing vehicle detection precision, thereby addressing limitations in earlier methods, such as poor generalization, appearance ambiguity, and fragmented tracking across multiple surveillance feeds.

## 5. Conclusion and Future Scope

This study developed an integrated framework for urban traffic management by combining CCTV surveillance with multi-source data fusion, frequent episode mining, and a hybrid LSTM–Transformer model. The approach addressed the limitations of traditional traffic systems by capturing spatio-temporal dependencies and sequential traffic behavior. Evaluation on the CityFlowV2 dataset demonstrated high performance, achieving 98.46% accuracy, with macro precision of 0.9800, macro recall of 0.9839, and macro F1-score of 0.9819. The mining of frequent traffic episodes revealed critical sequential patterns such as moderate–congested transitions. At the same time, the generation of 46 sustained congestion alerts confirmed the framework's ability to support proactive traffic management. These results highlight the significance of integrating video analytics with multi-source data for building adaptive and reliable transportation systems.

Future research can focus on reducing latency through edge computing and next-generation networks, incorporating crowdsourced and IoT-enabled vehicle data for richer context, and applying adaptive learning models that evolve with live conditions. Large-scale deployment across metropolitan areas and pilot testing in smart city environments can further validate scalability and real-world impact.

## Acknowledgments

The author would like to express her sincere gratitude to the members of "Aligarh Muslim University, Aligarh" for their invaluable contributions to this research work. Without their support, this study would not have been possible. The author is grateful to her coauthor for providing support, enthusiasm, and helpful criticism throughout the study process. The author emphasizes that no particular grant was awarded for this study by any public, private, or non-profit funding body. The costs associated with this study were borne entirely by the authors.

## Conflict of Interest

The authors state that there is no bias in their work.

# References

[1]. Paiva, Sara, Mohd Abdul Ahad, Gautami Tripathi, Noushaba Feroz, and Gabriella Casalino. "Enabling technologies for urban smart mobility: Recent trends, opportunities and challenges." *Sensors* 21, no. 6 (2021): 2143.

[2]. SUBAIR, Sulaiman Olayinka, Biliyamin Adeoye IBITOYE, and Abdulrauf Toyin KURANGA. "Evaluation of Traffic Congestion in an Urban Roads: A Review." *ABUAD Journal of Engineering and Applied Sciences* 2, no. 2 (2024): 1-7.

[3]. Moreno, Frede. "Traffic Congestion and Management in Zamboanga City, Philippines: The Public Transport Commuters' Point of View." *Philippines: The Public Transport Commuters' Point of View (July 14, 2023)* (2023).

[4]. Alharbi, Awad. "A framework for controlling & managing traffic with modern technology." PhD diss., City, University of London, 2021.

[5]. Agrahari, Anurag, Meera M. Dhabu, Parag S. Deshpande, Ashish Tiwari, Mogal Aftab Baig, and Ankush D. Sawarkar. "Artificial intelligence-based adaptive traffic signal control system: A comprehensive review." *Electronics* 13, no. 19 (2024): 3875.

[6]. Naveed, Quadri Noorulhasan, Hamed Alqahtani, Riaz Ullah Khan, Sultan Almakdi, Mohammed Alshehri, and Mohammed Aref Abdul Rasheed. "An intelligent traffic surveillance system using integrated wireless sensor network and improved phase timing optimization." *Sensors* 22, no. 9 (2022): 3333.

[7]. Nigam, Nikhil, Dhirendra Pratap Singh, and Jaytrilok Choudhary. "A review of different components of the intelligent traffic management system (ITMS)." *Symmetry* 15, no. 3 (2023): 583.

[8]. Nzesya, Luvai P. "Role of Technology in Crime Detection and Control in Urban Areas: the Case of Closed-circuit Television (Cctv) in Nairobi City." PhD diss., University of Nairobi, 2024.

[9]. Owais, Mahmoud, Ali Shehata, Abdou Shaband, and Ghada S. Moussa. "A Framework for Establishing an Automated Traffic Violation Detection System in New Assiut City Using Ordinary CCTV Units." *Mansoura Engineering Journal* 50, no. 3 (2025): 11.

[10]. Nallasivan, G. "A Vision-Based Traffic Accident Analysis and Tracking system from Traffic Surveillance Video." In *2024 Third International Conference on Intelligent Techniques in Control, Optimization and Signal Processing (INCOS)*, pp. 1-6. IEEE, 2024.

[11]. Siva, P., Garige Bhavani Pujitha, Gogula Siva Krishna, Gadde Hemanth, and Bonthagarla Manikanta Sai Teja. "Computer Vision Enabled Smart Surveillance for Urban Traffic Control." *International Journal on Advanced Computer Engineering and Communication Technology* 14, no. 1 (2025): 59-68.

[12]. Wong, Vivian WH, and Kincho H. Law. "Fusion of CCTV video and spatial information for automated crowd congestion monitoring in public urban spaces." *Algorithms* 16, no. 3 (2023): 154.

[13]. Tahir, Noor Ul Ain, Zuping Zhang, Muhammad Asim, Junhong Chen, and Mohammed ELAffendi. "Object detection in autonomous vehicles under adverse weather: A review of traditional and deep learning approaches." *Algorithms* 17, no. 3 (2024): 103.

[14]. Wang, Yizhe, Ruifa Luo, and Xiaoguang Yang. "Urban Traffic State Sensing and Analysis Based on ETC Data: A Survey." *Applied Sciences* 15, no. 12 (2025): 6863.

[15]. Devadhas Sujakumari, Praveen, and Paulraj Dassan. "Generative adversarial networks (GAN) and HDFS-based real-time traffic forecasting system using CCTV surveillance." *Symmetry* 15, no. 4 (2023): 779.

[16]. Okolo, Francess Chinyere, Emmanuel Augustine Etukudoh, Olufunmilayo Ogunwole, Grace Omotunde Osho, and Joseph Ozigi Basiru. "Advances in integrated geographic information systems and AI surveillance for real-time transportation threat monitoring." *Journal name missing* (2022).

[17]. Ounoughi, Chahinez. "Urban Traffic: Data Fusion and Vehicle Flow Prediction in Smart Cities."

[18]. Laanaoui, My Driss, Mohamed Lachgar, Hanine Mohamed, Hrimech Hamid, Santos Gracia Villar, and Imran Ashraf. "Enhancing urban traffic management through real-time anomaly detection and load balancing." *Ieee Access* 12 (2024): 63683-63700.

[19]. Hukasjan, Gor, and Ali Kazimov. "Data-Driven Traffic Forecasting in Södertälje Using LSTM Networks: Data-Driven Insights into Urban Mobility Patterns Using Machine Learning." (2025).

[20]. Khatri, Priti, Kaushlesh Singh Shakya, and Prashant Kumar. "A probabilistic framework for identifying anomalies in urban air quality data." *Environmental Science and Pollution Research* 31, no. 49 (2024): 59534-59570.

[21]. Tahir, Mehwish, Yuansong Qiao, Nadia Kanwal, Brian Lee, and Mamoona N. Asghar. "Real-time event-driven road traffic monitoring system using CCTV video analytics." *IEEE Access* 11 (2023): 139097-139111.

[22]. Ouarem, Oualid, Farid Nouioua, and Philippe Fournier-Viger. "A survey of episode mining." *Wiley Interdisciplinary Reviews: Data Mining and Knowledge Discovery* 14, no. 2 (2024): e1524.

[23]. Ferraro, Luca. "A parallel algorithm for mining sequences of spatio-temporal co-location patterns." PhD diss., Politecnico di Torino, 2023.

[24]. Lu, Jiaming, Chuanyang Hong, and Rui Wang. "MAGT-toll: A multi-agent reinforcement learning approach to dynamic traffic congestion pricing." *PloS one* 19, no. 11 (2024): e0313828.

[25]. Ashkanani, Manar, Alanoud AlAjmi, Aeshah Alhayyan, Zahraa Esmael, Mariam AlBedaiwi, and Muhammad Nadeem. "A Self-Adaptive Traffic Signal System Integrating Real-time Vehicle Detection and License Plate Recognition for Enhanced Traffic Management." Inventions 10, no. 1 (2025): 14.

[26]. Chaudhary, Ankur, M. Meenakshi, Soma Sharma, Mahbubur Rahman, and S. Srinivasan. "Enhancing urban mobility: machine learning-powered fusion approach for intelligent traffic congestion control in smart cities." International Journal of System Assurance Engineering and Management (2025): 1-8

[27]. Charef, Ayoub, Zahi Jarir, and Mohamed Quafafou. "Mobile Application Utilizing YOLOv8 for Real-time Urban Traffic Data Collection." In E3S Web of Conferences, vol. 601, p. 00077. EDP Sciences, 2025

[28]. Zahra, Asma, Mubeen Ghafoor, Kamran Munir, Ata Ullah, and Zain Ul Abideen. "Application of region-based video surveillance in smart cities using deep learning." Multimedia Tools and Applications 83, no. 5 (2024): 15313-15338

[29]. Li, Tao, Zilin Bian, Haozhe Lei, Fan Zuo, Ya-Ting Yang, Quanyan Zhu, Zhenning Li, and Kaan Ozbay. "Multi-level traffic-responsive tilt camera surveillance through predictive correlated online learning." Transportation Research Part C: Emerging Technologies 167 (2024): 104804

[30]. Devadhas Sujakumari, P., and P. Dassan. 2023. "Generative Adversarial Networks (GAN) and HDFS-Based Real-time Traffic Forecasting System Using CCTV Surveillance." Symmetry 15, no. 4: 779. https://doi.org/10.3390/sym15040779

[31]. Qin, J., J. Wang, Q. Li, S. Fang, X. Li, and L. Lei. 2022. "Differentially Private Frequent Episode Mining over Event Streams." Engineering Applications of Artificial Intelligence 110: 104681. https://doi.org/10.1016/j.engappai.2022.104681

[32]. Jiang, S. T., D. B. Zhou, Y. K. Chen, L. Gao, J. Z. Liu, and T. F. Li. 2022, Nov. "Holographic Traffic Signal Control System Based on Multi-source Data Fusion." In Seventh International Conference on Electromechanical Control. Technology and Transportation (ICECTT 2022) vol. 12302. SPIE. https://doi.org/10.1117/12.2645477

[33]. Li, Xiaohan, Jun Wang, Jinghua Tan, Shiyu Ji, and Huading Jia. 2022. "A Graph Neural Network-Based Stock Forecasting Method Utilizing Multi-source Heterogeneous Data Fusion." Multimedia Tools & Applications 81, no. 30: 43753–75. https://doi.org/10.1007/s11042-022-13231-1.

[34]. Zeng, X., X. Guan, H. Wu, and H. Xiao. 2021. "A Data-Driven Quasi-Dynamic Traffic Assignment Model Integrating Multi-source Traffic Sensor Data on the Expressway Network." ISPRS International Journal of Geo-Information 10, no. 3: 113. https://doi.org/10.3390/ijgi10030113

[35]. https://www.aicitychallenge.org/2022-data-and-evaluation/

[36]. Vrtagić, Sabahudin, Edis Softić, Mirza Ponjavić, Željko Stević, Marko Subotić, Aditya Gmanjunath, and Jasmin Kevric. "Video Data Extraction and Processing for Investigation of Vehicles' Impact on the Asphalt Deformation Through the Prism of Computational Algorithms." *Traitement du Signal* 37, no. 6 (2020).

[37]. Ullah, Fath U. Min, Mohammad S. Obaidat, Khan Muhammad, Amin Ullah, Sung Wook Baik, Fabio Cuzzolin, Joel JPC Rodrigues, and Victor Hugo C. de Albuquerque. "An intelligent system for complex violence pattern analysis and detection." International journal of intelligent systems 37, no. 12 (2022): 10400-10422.

[38]. Häb, Kathrin. "Visualization and Analysis Techniques for Urban Microclimate Data Sets." PhD diss., Technische Universität Kaiserslautern, 2015.

[39]. Li, Miaoyi. "The Possibility of Big Data Spatio-Temporal Analytics for Understanding Human Behavior and Their Spatial Patterns in Urban Area." PhD diss., Kanazawa University, 2017.

[40]. Chen, Hui, Jian Huang, Yong Lu, and Jijie Huang. "Urban traffic flow prediction based on multi-spatio-temporal feature fusion." *Neurocomputing* 638 (2025): 130117.

[41]. Zhou, Bodong, Jiahui Liu, Songyi Cui, and Yaping Zhao. "A large-scale spatio-temporal multimodal fusion framework for traffic prediction." *Big Data Mining and Analytics* 7, no. 3 (2024): 621-636.

[42]. Kou, Feifei, Ziyan Zhang, Yuhan Yao, Yuxian Zhu, Jiahao Wang, and Yifan Zhu. "A Survey on Long-Term Traffic Prediction from the Information Fusion Perspective: Requirements, Methods, Applications, and Outlooks." *Yifan, A Survey on Long-Term Traffic Prediction from the Information Fusion Perspective: Requirements, Methods, Applications, and Outlooks*.

[43]. Sarhan, Amany M. "Data mining in Internet of Things systems: A literature review." *Journal of Engineering Research* 6, no. 5 (2023): 252-263.

[44]. Mannila, Heikki, Hannu Toivonen, and A. Inkeri Verkamo. "Discovery of frequent episodes in event sequences." *Data mining and knowledge discovery* 1, no. 3 (1997): 259-289.

[45]. Guo, Youlin, and Zhizhong Mao. "Long-term prediction model for NOx emission based on LSTM–Transformer." *Electronics* 12, no. 18 (2023): 3929.

[46]. Chatterjee, Kaustav, Joshua Q. Li, Fatemeh Ansari, Masud Rana Munna, Kundan Parajulee, and Jared Schwennesen. "Hybrid LSTM-Transformer Models for Profiling Highway-Railway Grade Crossings." *arXiv preprint arXiv:2508.00039* (2025).

[47]. Yao, Hui, Zhizhao Duan, Zhen Xie, Jingbo Chen, Xi Wu, Duo Xu, and Yutao Gao. "City-scale multi-camera vehicle tracking based on space-time-appearance features." In *Proceedings of the IEEE/CVF conference on computer vision and pattern recognition*, pp. 3310-3318. 2022.

[48]. Li, Fei, Zhen Wang, Ding Nie, Shiyi Zhang, Xingqun Jiang, Xingxing Zhao, and Peng Hu. "Multi-camera vehicle tracking system for ai city challenge 2022." In *Proceedings of the IEEE/CVF Conference on Computer Vision and Pattern Recognition*, pp. 3265-3273. 2022.

[49]. Tang, Zheng, Milind Naphade, Ming-Yu Liu, Xiaodong Yang, Stan Birchfield, Shuo Wang, Ratnesh Kumar, David Anastasiu, and Jenq-Neng Hwang. "Cityflow: A city-scale benchmark for multi-target multi-camera vehicle tracking and re-identification." In *Proceedings of the IEEE/CVF conference on computer vision and pattern recognition*, pp. 8797-8806. 2019.

[50]. Hsu, Hung-Min, Yizhou Wang, Jiarui Cai, and Jenq-Neng Hwang. "Multi-target multi-camera tracking of vehicles by graph auto-encoder and self-supervised camera link model." In *Proceedings of the IEEE/CVF Winter Conference on Applications of Computer Vision*, pp. 489-499. 2022.

[51]. Lu, Jincheng, Xipeng Yang, Jin Ye, Yifu Zhang, Zhikang Zou, Wei Zhang, and Xiao Tan. "CityTrack: Improving city-scale multi-camera multi-target tracking by location-aware tracking and box-grained matching." *arXiv preprint arXiv:2307.02753* (2023).